\documentclass[conference]{IEEEtran}
\IEEEoverridecommandlockouts
\usepackage{cite}
\usepackage{amsmath,amssymb,amsfonts}
\usepackage{algorithmic}
\usepackage{graphicx}
\usepackage{textcomp}
\usepackage{multirow}
\usepackage{xcolor}
\usepackage{hyperref}
\def\BibTeX{{\rm B\kern-.05em{\sc i\kern-.025em b}\kern-.08em
    T\kern-.1667em\lower.7ex\hbox{E}\kern-.125emX}}
\begin{document}

\title{Sex-Prediction from Periocular Images across Multiple Sensors and Spectra}
\author{\IEEEauthorblockN{J. Tapia$^*$,  C. Rathgeb$^\dagger$ and C. Busch$^\dagger$}\vspace{0.2cm}
\IEEEauthorblockA{\textit{$^*$Universidad Tecnológica de Chile -- INACAP, Santiago, Chile} \\
\textit{$^\dagger$da/sec -- Biometrics and Internet Security Research Group, Hochschule Darmstadt, Germany}\vspace{0.2cm}\\
\texttt{j\_tapiaf@inacap.cl}\\
\texttt{\{christian.rathgeb,christoph.busch\}@h-da.de}}
\textbf{Pre-print version of Paper presented at Proc. International Workshop on Ubiquitous implicit Biometrics} \\ \textbf{ and health signals monitoring for person-centric applications (UBIO 18), 2018.}
}

\maketitle

\begin{abstract}

In this paper, we provide a comprehensive analysis of periocular-based sex-prediction (commonly referred to as gender classification) using state-of-the-art machine learning techniques. In order to reflect a more challenging scenario where periocular images are likely to be obtained from an unknown source, i.e. sensor, convolutional neural networks are trained on fused sets composed of several near-infrared (NIR) and visible wavelength (VW) image databases. In a cross-sensor scenario within each spectrum an average classification accuracy of approximately 85\% is achieved. When sex-prediction is performed across spectra an average classification accuracy of  about 82\% is obtained. Finally, a multi-spectral sex-prediction yields a classification accuracy of 83\% on average. Compared to proposed works, obtained results provide a more realistic estimation of the feasibility to predict a subject's sex from the periocular region. 
\end{abstract}


\section{Introduction}
\begin{figure}[t]
 \includegraphics[width=\linewidth]{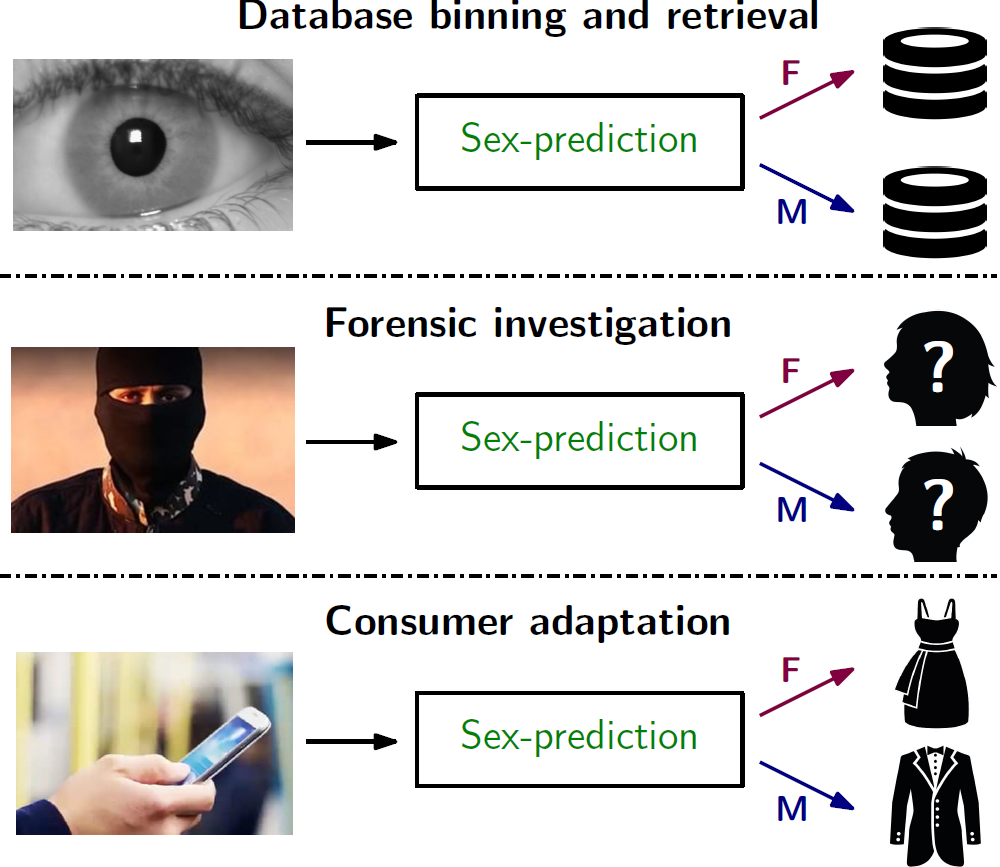}\label{fig:scenarios}\vspace{-0.2cm}
\caption{Different application scenarios where sex-prediction from  periocular images can be applied.}\vspace{-0.2cm}
\end{figure}

In the recent past, periocular biometrics, which refers to the externally visible skin region of the face that surrounds the eye socket, has been introduced for recognition purposes \cite{Fernandez16,Nigam15}. In many cases periocular recognition has been employed to augment the biometric performance of (VW) iris recognition in unconstrained environments, e.g. in mobile or surveillance scenarios. However, with the rise of deep learning in biometric technologies periocular recognition has gained practical biometric performance, too \cite{Proenca18,Zhao17}. Moreover, it has been demonstrated that the periocular region can be used to reliably predict soft biometric attributes \cite{Lyle10,Tapia_viedma2017}. Such attributes, e.g. an individual's sex, can be employed to glean demographic information or improve or expedite recognition capabilities in conjunction with primary biometric characteristics \cite{Dantcheva16}. In particular, sex-prediction turns out to be a pre-processing step in a biometric system with many applications ranging from forensic analysis to database binning, see Fig. \ref{fig:scenarios}. 

In past years, researchers have demonstrated that diverse soft-biometric attributes can be obtained from (peri-)ocular images. Eye colour which can be naturally gleaned from an iris images acquired in visible band of light has been utilized as additional source of information in different approaches, e.g. in \cite{BTan12a,Dantcheva11a}. The existence of further soft biometric features in the iris pattern, such as a subject's sex, age or ethnicity, has been analyzed in various works \cite{Lagree11a,Fairhurst15a,Bobeldyk16a,TapiaAravena2017,SinghNagpalVatsaEtAl2017}. However, it is generally conceded that these attributes can be extracted more reliably if  parts of the periocular region are additionally available \cite{Merkow10,Bobeldyk16a,KuehlkampBeckerBowyer2017,Tapia_Aravena2018}.

\begin{table*}[t]
\small
\centering
\caption{Most relevant approaches to sex-prediction from periocular images.}\vspace{0.0cm}
\label{tab:related}
\begin{tabular}{|c|c|c|c|c|c|}
\hline
\textbf{Authors} & \textbf{Approach} & \textbf{Database} & \textbf{Spectrum} & \textbf{Accuracy} & \textbf{Remarks} \\ \hline
Merkow \textit{et al.} \cite{Merkow10} & LBP with SVM & retrieved from Flickr (936 imgs.) & VW & 79.6\% & private DB\\ \hline
Bobeldyk and Ross \cite{Bobeldyk16a} & BSIF with SVM & BioCOP (1,720 imgs.) & NIR & 85.7\% & private DB \\ \hline 
Tapia and Aravena \cite{Tapia_Aravena2018} & CNN & GFI\_UND \cite{TapiaPerezBowyer2016} & NIR & 89.0\% & -- \\\hline
Kuehlkamp \textit{et al.} \cite{KuehlkampBeckerBowyer2017} & CNN & GFI\_UND \cite{TapiaPerezBowyer2016} & NIR & 66.0\% & -- \\\hline
Ours & CNN & 10 public DBs ($>$15,000 imgs.) & VW and NIR & 86.6\% & -- \\\hline
\end{tabular}
\end{table*}

\begin{figure*}[t]
\centering
  \includegraphics[height=4.65cm]{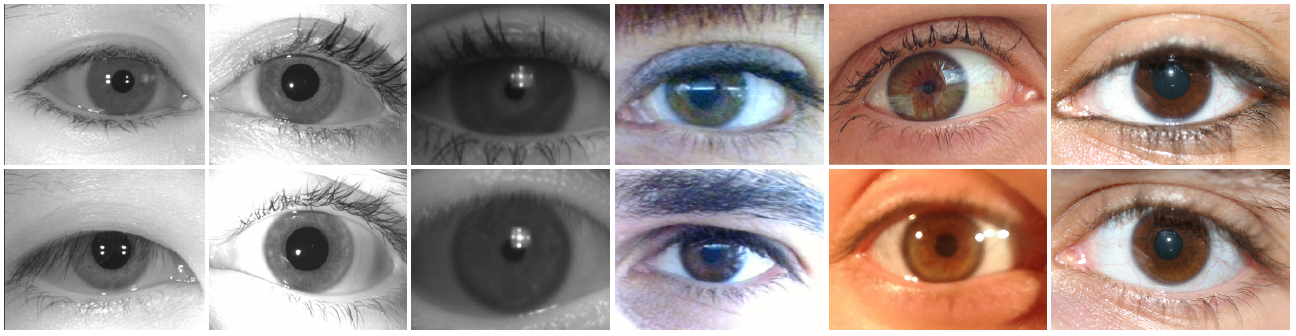}
\caption{Examples of female (top) and male (bottom) NIR and VW ocular images of different employed databases; from left to right: CASIA-DISTANCE, GFI-UND, CROSS-EYED, MOBBIO, MICHE, UTIRIS. }\label{fig:examples}\vspace{-0.2cm}
\end{figure*} 

Focusing on sex-prediction from periocular biometrics the vast majority of proposed approaches combines general purpose texture descriptors such as local binary patterns in conjunction with machine learning-based classifiers, e.g. \cite{Merkow10,Bobeldyk16a}. Moreover, the application of deep neural networks for the task of sex-prediction from (peri-)ocular images has been investigated recently \cite{KuehlkampBeckerBowyer2017,Tapia_Aravena2018}. Even for the binary classification problem of sex-prediction deep neural networks require a sufficient amount of labeled training images to achieve practical classification accuracy. However, the amount of publicly available (peri-)ocular image datasets which comprise sex labels is very limited, i.e. in most cases to rather unbalanced sets of less than a hundred female and male subjects. While promising classification accuracy has been reported for the task of sex-prediction from the periocular region,  existing studies are primarily conducted on datasets of distinct format acquired with a single sensor, see Table \ref{tab:related}. That is, related works restrict to scenarios in which system is presented with images from a single source. This means, information such as image type, format, or sensor are assumed to be known and according training data is used. 

The increasing popularity of (peri-)ocular biometrics leads to a continuous development and upgrade of sensors, respectively. Differences among multiple types of  sensors such as optical lens and illumination wavelength yield certain cross-sensor variations which might lead to reduced biometric performance \cite{Connaughton12a,Arora12a}. A re-enrolment of subjects every time a new sensor is deployed is expensive and time-consuming, especially in large-scale applications.
This fact motivated the proposal of diverse approaches which aim at improving the biometric performance of cross-sensor ocular recognition systems, e.g. in \cite{Xiao13a,Pillai14a}. In addition, the case in which the spectrum of the image acquisition changes, i.e. from NIR to VW and vice versa, has been considered, e.g. in  \cite{Jillela14a,Sequeira16a}. It was found that the recognition accuracy significantly decreases in such a scenario. Regarding cross-spectral soft biometric feature extraction based on (peri-)ocular images, it was recently shown that eye colour might be predicted from NIR ocular images \cite{Bobeldyk18a}.

\begin{table*}[t]
\small
\centering
\caption{NIR databases: \emph{F} represents the number of Female images and \emph{M} the number of Male images.}\vspace{0.0cm}
\label{tab_nir}
\begin{tabular}{|c|c|c|c|c|c|c|}
\hline
\textbf{Dataset} & \textbf{Resolution} & \textbf{No. Images} & \textbf{No. Subjects} & \textbf{F} & \textbf{M} & \textbf{Sensor}\\ \hline
CASIA-DISTANCE \cite{TAN2010223} & 250$\times$200 (cropped)  & 2,567  & 141 & 53 & 88  & LMBS  \\ \hline
 CROSS-EYED \cite{Sequeira16a}   & 400$\times$300      & 1,920  & 120     & 41   & 79  & n.a.  \\ \hline
GFI\_UND \cite{TapiaPerezBowyer2016}& 640$\times$480     & 3,000  & 1,500   & 750  & 750  & LG 4000       \\ \hline
UND\_VAL \cite{TapiaPerezBowyer2016}& 640$\times$480 & 1,944 & 324  & 149   & 175  & LG 4000          \\ \hline
UTIRIS \cite{utiris2010} & 1,000$\times$776  & 389 & 79  & 13   & 66   & 
ISG LW\\ \hline
\end{tabular}\vspace{-0.2cm}
\end{table*}

\begin{table*}[t]
\small
\centering
\caption{VW databases:  \emph{F} represents the number of Female images and \emph{M} the number of Male images; (*) only left images available.}\vspace{0.0cm}
\label{tab_vis}
\begin{tabular}{|c|c|c|c|l|l|c|}
\hline
\textbf{Dataset} & \multicolumn{1}{l|}{\textbf{Resolution}}                                                                                           & \multicolumn{1}{l|}{\textbf{No. Images}} & \multicolumn{1}{l|}{\textbf{No. Subjects}} & \textbf{F} & \textbf{M} & \textbf{Sensor(s)}                                                                              \\ \hline
CROSS-EYED \cite{Sequeira16a} & 400$\times$300  & 1,920 & 120  & 41 & 79         & n.a.    \\ \hline
UTIRIS \cite{utiris2010}  & 1,000$\times$776 & 389 & 79 & 13     & 66         & Canon EOS 10D                                  \\ \hline
CSIP(*) \cite{SANTOS201552}   & var. res. & 2,004 & 50                                     & 9          & 41         & \begin{tabular}[c]{@{}c@{}}Xperia ArcS, iPhone 4,\\ Th.I W200, Hua U8510\\ \end{tabular} \\ \hline
MOBBIO \cite{Sequeira14a}  & 250$\times$200  & 800 & 100 & 29  & 71  & Asus Eee Pad Transformer TE300T \\ \hline
MICHE \cite{DeMarsico2015}   & 1,000$\times$776 & 2,732 & 92   & 26 & 76 & iPhone 5 (subset) \\ \hline
\end{tabular}\vspace{-0.2cm}
\end{table*}

In this work, sex-prediction from periocular images is performed employing convolutional neural networks (CNNs) which represent the current state-of-the-art for the extraction of soft-biometric features \cite{Dantcheva16}. In order to obtain labeled sex-balanced training datasets of sufficient size  multiple publicly available NIR and VW image databases are fused. At the same time, training on multiple images sources, acquired in various formats using different NIR and VW sensors, allows for a deeper analysis with respect to robustness in a cross-sensor scenario within the NIR and VW spectrum. In addition, the generalizability of the proposed sex-prediction systems is investigated in a cross-spectral scenario, which means that the training procedures are performed on fused datasets of NIR and/or VW images and evaluations are performed on VW and/or NIR images, respectively. Hence, we perform sex-prediction based on (peri-)ocular images considering  challenging classification scenarios across multiple sensors and spectra employing state-of-the-art machine learning techniques. In contrast to existing studies, which suffer from aforementioned limitation, the proposed study reflects scenarios where the image source of a (peri-)ocular images might be unknown to system, which can be the case in a real-world scenario. To the authors' knowledge there exist no works on \emph{cross-sensor}, \emph{cross-spectral} or \emph{multi-spectral} sex-prediction from (peri-)ocular images.

This paper is organized as follows:  The databases used in this study are summarized in detail in Sect. \ref{sec:setup}. Employed CNN-based sex-prediction methods are described in Sect. \ref{sec:system}. Obtained results are reported and discussed in Sect. \ref{sec:experiments}. Finally, conclusions are drawn in Sect. \ref{sec:conclusion}.

\section{Databases}\label{sec:setup}

Generally speaking, publicly available (peri-)ocular biometric databases were mostly created to facilitate the development of recognition algorithm. Hence, said VW and NIR databases commonly comprise multiple ocular images per data subject acquired within one or across several sessions. Unfortunately, many available databases do not provide soft-biometric attributes like the sex of data subjects. Due to this reason, the creation of a labeled large-scale ocular image database including sex information, which is necessary to properly train machine learning-based classifiers, represents a challenging tasks.

The databases used in this study are listed in Table \ref{tab_nir} for NIR spectrum and Table \ref{tab_vis} for VW, respectively. Sample images of NIR as well as VW ocular images are shown in Fig. \ref{fig:examples}. Note, that the amount of periocular information present in the processed images may vary. Also the spectral band may vary for NIR images. Since  images are directly used for the task of sex-prediction all images are referred to as periocular images. While NIR images are mostly acquired under more constrained conditions, VW images show higher variations in the capture process. For some of the employed databases, e.g. GFI\_UND or UND\_VAL, sex information is available. For others, e.g. CROSS-EYED or UTIRIS, sex information is available upon request. Eventually, remaining databases, e.g. CASIA-DISTANCE or MOBBIO, were manually labeled based on (corresponding) face images. In case only face images were available within databases, e.g. CASIA-DISTANCE or MICHE (iPhone 5), the OpenCV 2.10 eye detector was employed to automatically detect and crop the left and right ocular regions. Images where the eye detector failed were deleted from the database.  Another alternative to obtain a sufficiently large database of periocular images with sex-labels would be to process further face databases for which these labels are available.

In order to provide a uniform (and compact) image format for the sex-prediction algorithm during training, all images (NIR and VW) are cropped and re-scaled to the minimum available image resolution, i.e. 250$\times$200 pixels for the MOBBIO database (see Table \ref{tab_vis}). Cropping is performed to retain the aspect ratio of each image, such that stretching of images is prevented. Finally, VW images are converted to grayscale. On the one hand, a simple grayscale conversion is applied, on the other hand, the red channel is extracted. The latter method is motivated by the fact that the red color spectrum is nearer to the NIR spectrum. That is, the use of red channel information only is expected to facilitate the cross-spectral performance of a sex-prediction algorithm.

\section{CNN-based Sex-Prediction}\label{sec:system}

Three different ways of how to obtain a CNN-based sex-prediction algorithm are described in the following subsections\footnote{details of CNN-based classifiers are summarized according to the guidelines provided by the IEEE Signal Processing Society}: (1) training a CNN from scratch, (2) using the so-called bottleneck features of a pre-trained network and (3) fine-tuning the top layers of a pre-trained network. The latter two approaches represent instances of \emph{transfer learning}  \cite{Yosinski2014}. In transfer learning, a base network is trained on a base dataset and task, and then the learned features are reused or transferred to a second target network to be trained on a target dataset and task. This process is expected to work if the features learned on the base dataset are general, meaning suitable to both, base and target tasks.

\subsection{Training from Scratch}
Firstly, CNNs are trained with different filter sizes from scratch using a small learning rate of $10^{-5}$, such that each new set of fully connected layers can start learning patterns from the previously learned convolutional layers at an early point. The weights were randomly initialized with $100$ epochs. Optionally, the rest of the network might be unfrozen before continuing training. The resulting small CNN network can be seen as baseline system.

In order to obtain an optimal set of network parameters a large amount of training data is needed for this approach. Moreover, when training CNNs from scratch a huge of computing resources might be required.   

\subsection{Transfer Learning}

When the target dataset is significantly smaller than the base dataset, transfer learning can be a powerful tool to enable training a large target network without overfitting. Recent studies have taken advantage of this fact to obtain state-of-the-art results when transferring from higher layers \cite{Yosinski2014}, collectively suggesting that these layers of neural networks do indeed compute features that are fairly general.

Training data of another domain might be in a different feature space or follow a different data distribution. However, the amount of training data is sufficiently large and the resulting network can be directly used or fine-tuned for the task of sex-prediction from periocular images.

\subsubsection{Bottleneck features}
Within this approach the bottleneck features of a pre-trained network (VGG-19) are leveraged. This approach is motivated by the fact that such a pre-trained network is expected to have learned features that are useful for diverse pattern recognition problems. In particular, the VGG-19 architecture is trained on the ImageNet dataset which contains a total number of merely 1,000 different classes. Hence, these models are not expected to have learned (peri-)ocular features that are relevant to the sex-prediction problem. In fact, it is not possible to merely record the soft-max predictions of the model over the analyzed data rather than the bottleneck features in order to solve the said classification problem. However, a model trained on a large dataset will contain many learned basic features like edges, spots, ridges or horizontal lines that might be transferable to the periocular datasets. Only the convolutional part of the model is instantiated, i.e. everything up to the fully-connected layers. Subsequently, the model is run on our training and test data once, recording the output  in two arrays, i.e. the "bottleneck features" from the VGG-19 model (the last activation maps before the fully-connected layers). Then we trained a small fully-connected model on top of the stored features.
The reason why we are storing the features off-line rather than adding our fully-connected model directly on top of a frozen convolutional base, is computational efficiency. Obviously, the usefulness of employed bottleneck features will highly depend on their generality. Hence, this approach will work in case the features required for robust sex-prediction from periocular images are fairly general.

\subsubsection{Fine-tuning}
Finally, fine-tuning allows to apply pre-trained networks to recognize classes that they were not originally trained on. This method can lead higher accuracy compared to regular feature extraction methods. Therefore, the last convolutional block of the VGG-19 model is ``fine-tuned'' alongside the top-level classifier \cite{SimonyanZ14a}.   In the proposed study, fine-tuning, which represents a type of transfer learning, is performed in three steps: (a) instantiate the convolutional base of VGG-19 and load its weights; (b) add the previously defined fully-connected model on top, and load its weights; (c) freeze the layers of the VGG-19 model up to the last convolutional block. Fine-tuning is done with a very slow learning rate, too. The SGD optimizer is used rather than an adaptive learning rate optimizer, e.g. RMSProp. This is done to make sure that the magnitude of the updates remains very small, so not to suppress the previously learned features. The learning rate of the SGD value was set-up to $10^{-5}$ and the momentum of $0.8$.

In summary, fine-tuning aims at avoiding the aforementioned limitations of insufficient amount training data in case of training a CNN from scratch and the incapacity of features of pre-trained CNNs.

\section{Experiments}\label{sec:experiments}

In the following subsections, the experimental protocol, the applied data augmentation and the proposed CNN-based sex-prediction scheme are described in detail. Subsequently, obtained results are reported and discussed. All experiments were conducted using the Python open-source libraries Theano\footnote{\url{http://deeplearning.net/software/theano/}} and Keras\footnote{\url{https://keras.io/backend/}}.%

\subsection{Evaluation Protocol}

As previously mentioned, the proposed study aims at analyzing the generalizability of state-of-the-art algorithms for sex-prediction from periocular images. We investigate the following scenarios:
\begin{itemize}
    \item \emph{Cross-sensor}: sex-prediction algorithms are trained on fused datasets of NIR or VW ocular images acquired by different types of sensors in various formats. Subsequently, testing is performed within the same spectrum. 
    \item \emph{Cross-spectral}: sex-prediction algorithms trained in the above scenario are evaluated using test images of the other spectrum, i.e. algorithms trained on NIR images are tested on VW images and vice versa. The cross-spectral scenario might be seen as a special cross-sensor case in which the type of sensor changes. 
    \item \emph{Multi-spectral}: sex-prediction algorithms are trained on a fused dataset of NIR and VW ocular images acquired by different types of sensors in various formats. Accordingly, the test set comprises images of both spectra.
\end{itemize}

Moreover, it is important to note the partitioning of training and test sets. To  achieve  a  fair  and  meaningful  comparison between the tested methodologies, each  NIR and VW  is divided as follows: for each employed database the images of 50\%  of female subjects and the same amount of male subjects were selected for model  training. Due to the fact that all used databases comprise more male than female subjects (M$>$F), a balanced training set is pre-selected. The remaining images are used for the test set. Members of subsets were randomly selected.

The described procedure leads to a training set of 2,025 female and male NIR periocular images and a test set of 2,004 female and 4,035 male NIR periocular images. Similarly, a training set of 1,092 female and male VW periocular images and a test set of 1,017 female and 4,235 male images is obtained. The same partitioning of datasets is applied in experiments where only the red color channel is used during grayscale conversion.

\begin{table*}[t]
\centering
\small
\caption{Performance of the sex-prediction algorithms in diverse scenarios: \emph{ACC} represent accuracy of sex-prediction trained from scratch; \emph{ACC-DA} represent accuracy with data-augmentation trained from scratch; \emph{RED} refers to VW images where the red channel is extracted; two rightmost columns summarize transfer learning approaches.}\label{results}\vspace{0.0cm}
\label{my-label}
\begin{tabular}{|c|c|c|c|c|c|c|}
\hline
\multirow{2}{*}{\textbf{\begin{tabular}[c]{@{}c@{}}Scenario  \end{tabular}}} & \multicolumn{2}{|c|}{\textbf{Dataset}} & \multirow{2}{*}{\textbf{\begin{tabular}[c]{@{}c@{}}ACC  (\%)\end{tabular}}} & \multirow{2}{*}{\textbf{\begin{tabular}[c]{@{}c@{}}ACC-DA  (\%)\end{tabular}}} & \multicolumn{2}{|c|}{\textbf{VGG-19}}  \\ \cline{2-3} \cline{6-7} 
& \textbf{Training}     & \textbf{Test}     &                                                                              &                                                                                 & \textbf{Bottleneck (\%)}  & \textbf{Fine-tuning (\%)}       \\ \hline 
\multirow{3}{*}{\begin{tabular}[c]{@{}c@{}}Cross-sensor  \end{tabular}} & NIR                & NIR               & 67.55                                                                        & 71.95                                                                           & 78.40                                                           & 80.90                                                                \\ \cline{2-7}
& VW                & VW               & 68.15                                                                        & 72.05                                                                           & 77.65                                                           & 85.90                                                                \\ \cline{2-7}
& RED                & RED               & 69.75                                                                        & 77.59                                                                           & 83.40                                                           & \textbf{88.90}                                                                \\ \hline
\multirow{6}{*}{\begin{tabular}[c]{@{}c@{}}Cross-spectral  \end{tabular}} & NIR                & VW               & 68.50                                                                        & 71.80                                                                           & 75.60                                                           & 78.30                                                                \\ \cline{2-7}
& VW                & NIR               & 68.65                                                                        & 72.35                                                                           & 70.25                                                           & 72.25                                                                \\ \cline{2-7}

& NIR                & RED               & 65.60                                                                        & 73.45                                                                           & \textbf{85.45}                                                           & \textbf{91.90}                                                                \\ \cline{2-7}
& RED                & NIR               & 65.50                                                                        & 67.25                                                                           & 71.50                                                           & 79.90                                                                \\ \cline{2-7}
& VW                & RED               & 65.70                                                                        & 71,25                                                                          & 71.45                                                          & 70.85                                                               \\ \cline{2-7}
& RED                & VW               & 68.45                                                                        & 72.15                                                                           & 75.50                                                           & 87.45                                                                \\ \hline
\multirow{4}{*}{\begin{tabular}[c]{@{}c@{}}Multi-spectral  \end{tabular}} & NIR-VW            & NIR-VW           & 68.25                                                                        & 71.45                                                                           & 73.45                                                           & 77.95                                                                \\ \cline{2-7}
& NIR-RED            & NIR-RED          & 68.50                                                                        & 72.45                                                                           & 81.30                                                           & 85.30                                                                \\ \cline{2-7}
& VW-RED            & VW-RED           & 70.65                                                                        & 75.35                                                                           & 79.60                                                           & 81.65                                                                \\ \cline{2-7}
& VW-RED-NIR            & VW-RED-NIR           & \textbf{74.65}                                                                        & \textbf{79.15}                                                                           & 84.25                                                           & \textbf{86.60}                                                                \\ \hline
\end{tabular}\vspace{-0.2cm}
\end{table*}

\begin{figure}[t]
\begin{centering}
\includegraphics[scale=0.25]{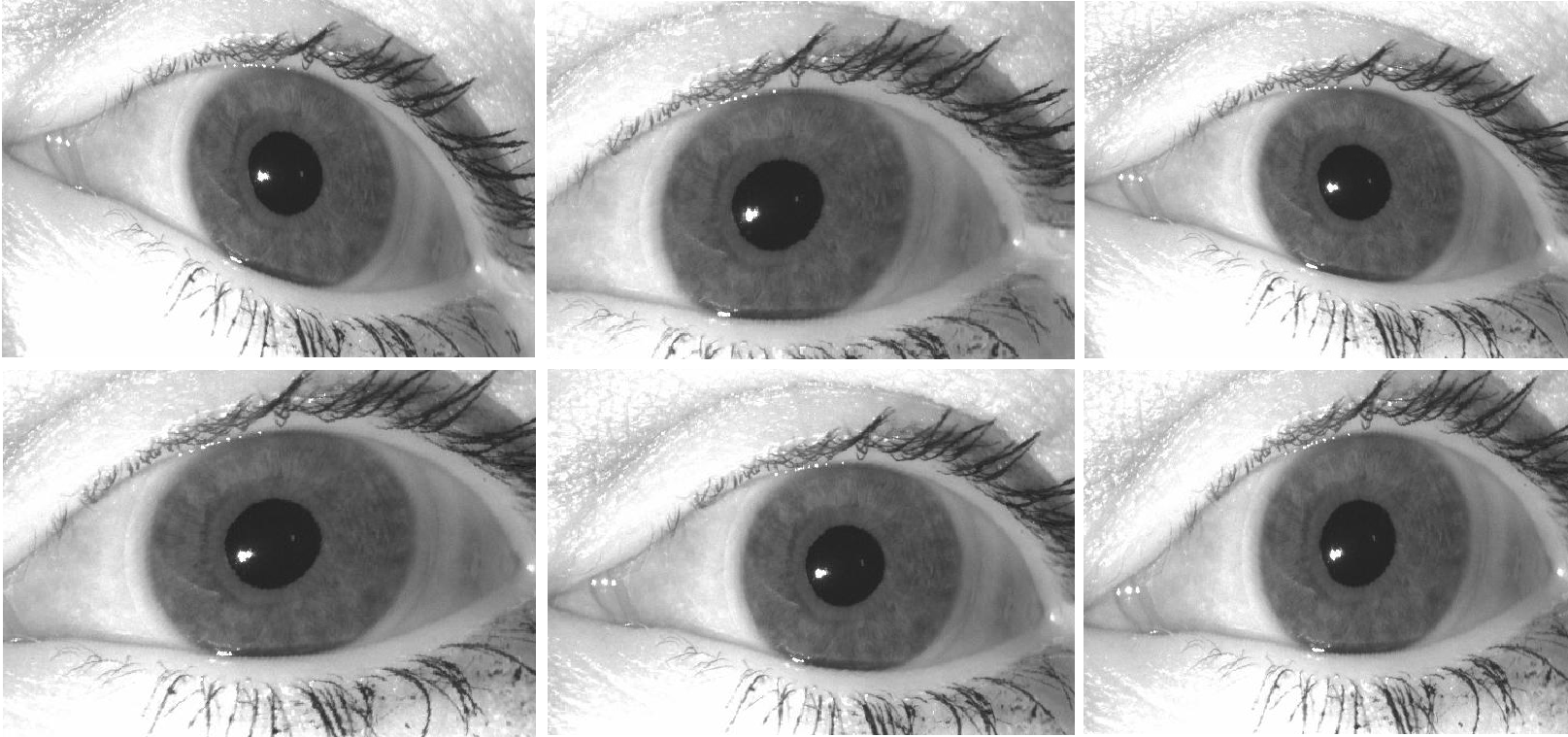}
\par\end{centering}
\caption{Examples images resulting from the data augmentation process used in the training stage of the CNN models.}\label{fig:da}\vspace{-0.2cm}
\end{figure}
\subsection{Data Augmentation}

To artificially increase the number of training images, and thus the robustness of the developed CNN models, an image generator function was used. With respect to the investigated classification problem the following geometric transformations have been identified as most relevant: image rotation within a range of 5 degree, image shifts with range of 20\% and a zoom range within 10\%. All image modification are performed using a 'Nearest' fill mode. Furthermore, mirroring was applied since both, left and the right peri-ocular images to extend the training sets. The process of data augmentation is shown in Fig. \ref{fig:da} for a single NIR ocular image. By applying data augmentation a fix-fold increase of all datasets is achieved.

\subsection{Sex-Prediction from Periocular Images}

In the proposed sex-prediction algorithm CNN-based models are trained with $100$ epochs using the network architecture shown in Fig. \ref{fig:cnn}. The number of images available to train a CNN from scratch is limited and in order not to overtrain a low number of layers based on Lenet-5 was chosen. The number of layers (from 2 up to 5) and a trade-off with the number of filters and sizes of the kernels were analyzed. The results reported in the paper represent the best parameters for the classification task. The four subsequent convolutional layers are defined as follows:

\begin{enumerate}
\item Conv1 filters of size $1\times10\times10$ pixels are applied to the input in the first convolutional layer, followed by a rectified linear operator (ReLU), a max pooling layer taking the mean value of $2\times2$ regions with two-pixel strides and a local response normalization layer.
\item Conv2 filters of size $1\times15\times15$ pixels are applied to the input in the second convolutional layer, followed by a rectified linear operator (ReLU), a max pooling layer taking the mean value of $2\times2$ regions with two-pixel strides and a local response normalization layer.
\item Conv3 filters of size $1\times20\times20$ pixels are applied to the input in the third convolutional layer, followed by a rectified linear operator (ReLU), a max-pooling layer taking the mean value of $2\times2$ regions with two-pixel strides and a local response normalization layer.

\begin{figure}[t]
\begin{centering}
\includegraphics[scale=0.285]{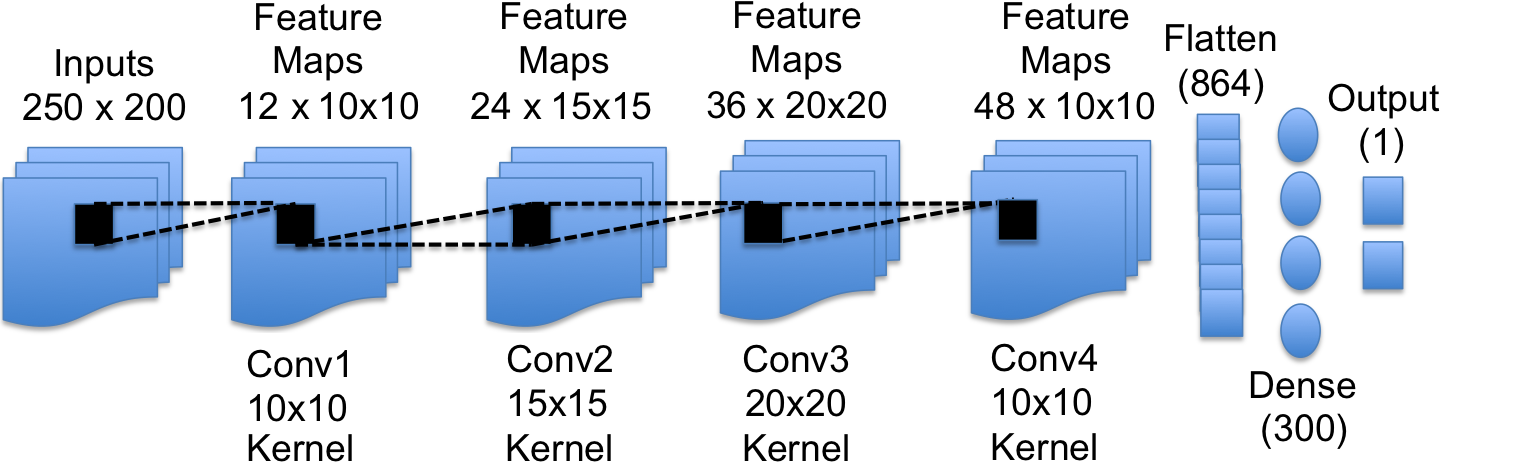}
\par\end{centering}
\caption{Architecture of employed CNN comprising four convolutional layers.}\label{fig:cnn}\vspace{-0.2cm}
\end{figure}

\item Conv4 filters of size $1\times10\times10$ pixels are applied to the input in the fourth convolutional layer, followed by a ReLU operator, a max pooling layer taking the mean value of $2\times2$ regions with two-pixel strides and a local response normalization layer.

A final connected layer is then defined by:

\item A flatten connected layer that receives the output of the fourth convolutional with 864 neurons and a Dense layer that contains 300 neurons followed by ReLU.

\item Finally, the output of the last Dense  layer is fed to a sigmoid layer that assigns a probability for each class. The prediction itself was made by taking the class with the maximal probability for the given test image.
\end{enumerate}

\begin{figure*}[t]
\centering
  \includegraphics[height=4.9cm]{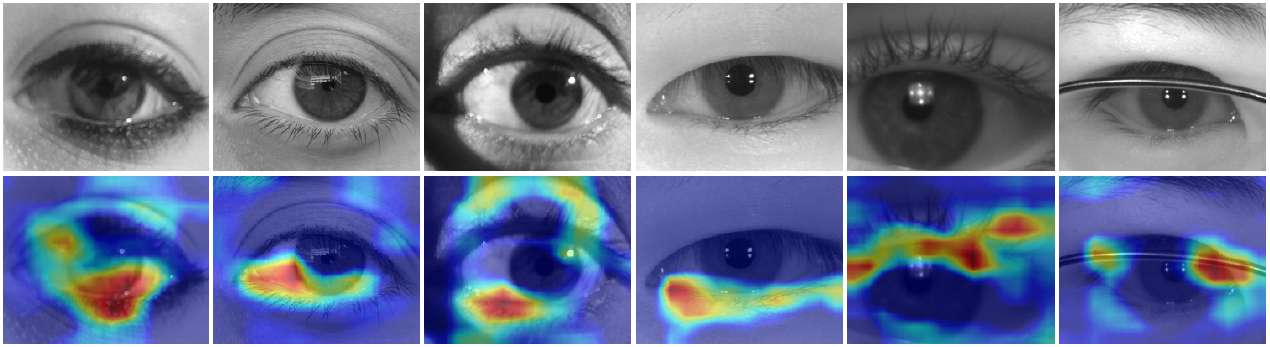}\vspace{-0.1cm}
\caption{Examples of female and male images acquired at NIR and VW and the resulting activation maps obtained for the best CNN-based sex-prediction method; warmer colors (red) represent higher responses, i.e. pixels at these positions are more important for sex-prediction; the rightmost image shows an example for miss-classification due to glass frames.}\label{fig:activation}\vspace{-0.3cm}
\end{figure*} 

\subsection{Performance Evaluation}

Table \ref{results} summarizes obtained results in terms of correct classification rate for the CNN-based sex-prediction algorithms in various scenarios. Best obtained performance for distinct scenarios and training techniques are marked bold. As can be seen, classification accuracies are similar to those reported in related works, cf. Table \ref{tab:related}. Interestingly, no significant decrease in classification accuracy can be observed between the \emph{cross-sensor} and \emph{cross-spectral} scenario (approx. 65-70\%), where the latter is generally conceded as more challenging. This could imply that sex information is similar across spectra. Further, competitive results are obtained in the \emph{multi-spectral} scenario, in particular for VW-RED-NIR, where all training sets are combined. This can be explained by the increase in training data. That is, sex information does not seem to be more pronounced in a single spectrum. Moreover, it can be observed that for each scenario the best results are mostly achieved if the training is performed on NIR images. Compared to VW images, NIR images are acquired under more constrained conditions which can be the reason for this performance gain.

The performance of the CNN which has been trained from scratch is generally inferior compared to the remaining training techniques, since the number of labeled images available was small. Hence, even when an increased number of convolutional layers is used classification accuracy is not improved.  Data-augmentation improved the performance in all scenarios even though the degree of data-augmentation was kept small in order not over-trained the CNN models. Focusing on the use of bottleneck feature from a pre-trained network for sex prediction, in the \emph{cross-sensor} scenario result can be significantly improved (approx. 77-84\%). Similar improvements are achieved in the  \emph{multi-spectral} scenario and for individual configurations of the \emph{cross-spectral} scenario. Fine tuning further enhances the classification performance achieving best sex-prediction accuracy in all considered scenarios. Obtained results suggest, that \emph{cross-sensor}, \emph{cross-spectral} and \emph{multi-spectral} sex-prediction is feasible. 
As expected, an increased training set leads to improved classification accuracy, which is observed for the \emph{multi-spectral} scenario.

In order to show the activation maps of the best model of VGG-19 with the transfer learning approach, we used the gram-cam visualization proposed in \cite{Selvaraju_2017_ICCV}. Examples of VW and NIR images and the resulting activation maps are shown in Fig. \ref{fig:activation}. It is found that activation maps are very similar for NIR and VW images, especially for VW and RED the activation maps are almost identical. This further supports the claim that sex-related information is similar across different spectra. It can be observed that for the depicted examples upper and lower eyelid parts carry most important features for sex-prediction. Obviously, the presence of mascara, eyeshadow or eyeliner in those parts of the periocular region is a strong indicator for an image to stem from a female subject. Similarly, images containing thick eyebrows are more likely to stem from male subjects. Such soft-biometric features within the periocular region are vital for robust sex-prediction. Nevertheless, it is important to note that subjects can take influence on said features and classification performance is expected to drop if these are absent \cite{KuehlkampBeckerBowyer2017}.

\section{Conclusion}\label{sec:conclusion}

In many biometric research areas the application of deep neural networks has revealed significant improvements. This work analyses the generalizability of CNN-based algorithms for sex-prediction from periocular images. By analyzing various realistic application scenarios (as apposed to many published approaches) it is shown that CNN-based sex-prediction algorithms are capable of classifying periocular images in \emph{cross-sensor}, \emph{cross-spectral} and \emph{multi-spectral} scenarios where the application of data augmentation and transfer learning reveals competitive classification accuracy. Taking into account that in many use-cases the left and right periocular region of a data subjects will be available, the robustness of CNN-based sex-prediction algorithms suggests that the periocular region is a suitable biometric source for predicting a subject's sex. 

\section*{Acknowledgments}
This work was partially supported by the German Federal Ministry of Education and Research (BMBF), the Hessen State Ministry for Higher Education, Research and the Arts (HMWK) within the Center for Research in Security and Privacy (CRISP) and by Fondecyt Iniciacion N11170189 at Universidad Tecnológica de Chile (INACAP).
\bibliographystyle{IEEEtran}
\bibliography{IEEEabrv}

\end{document}